\documentclass[10pt,twocolumn,letterpaper]{article}

\usepackage[pagenumbers]{cvpr} 

\usepackage{graphicx}
\usepackage{amsmath}
\usepackage{amssymb}
\usepackage{amsfonts,amsthm}
\usepackage{booktabs}
\usepackage{subcaption}
\usepackage{xcolor}

\usepackage[pagebackref,breaklinks,colorlinks]{hyperref}

\usepackage[capitalize]{cleveref}
\crefname{section}{Sec.}{Secs.}
\Crefname{section}{Section}{Sections}
\Crefname{table}{Table}{Tables}
\crefname{table}{Tab.}{Tabs.}

\newcommand{\algname}{\mbox{ScePT}}

\def\cvprPaperID{9886} 
\def\confName{CVPR}
\def\confYear{2022}

\author{Authors}

\begin{document}

\title{
\algname{}: Scene-consistent, Policy-based Trajectory Predictions for Planning}

\author{Yuxiao Chen$^1$ \hspace{1cm} Boris Ivanovic$^1$ \hspace{1cm} Marco Pavone$^{1,2}$\\
$^1$NVIDIA Research \hspace{1cm} $^2$Stanford University\\
{\tt\small \{yuxiaoc, bivanovic, mpavone\}@nvidia.com, pavone@stanford.edu}
}
\maketitle

\begin{abstract}
Trajectory prediction is a critical functionality of autonomous systems that share environments with uncontrolled agents, one prominent example being self-driving vehicles. Currently, most prediction methods do not enforce scene consistency, i.e., there are a substantial amount of self-collisions between predicted trajectories of different agents in the scene. Moreover, many approaches generate individual trajectory predictions per agent instead of joint trajectory predictions of the whole scene, which makes downstream planning difficult. In this work, we present \algname, a policy planning-based trajectory prediction model that generates accurate, scene-consistent trajectory predictions suitable for autonomous system motion planning. It explicitly enforces scene consistency and learns an agent interaction policy that can be used for conditional prediction. Experiments on multiple real-world pedestrians and autonomous vehicle datasets show that \algname{} matches current state-of-the-art prediction accuracy with significantly improved scene consistency. We also demonstrate \algname's ability to work with a downstream contingency planner.
\end{abstract}

\section{Introduction}
Predicting the future motion of uncontrolled agents is critical to the safety of autonomous systems that interact with them. A prominent example is self-driving cars, where the ego-vehicle shares the road with other road users such as vehicles, pedestrians, and cyclists. The prediction task is difficult as humans are notoriously uncertain and inconsistent. For example, it is well-known that humans demonstrate multimodal behaviors in the context of driving, such as being simultaneously able to maintain their current lane, change lanes, yield, or overtake in the future. As a result, early works on human driving behavior prediction~\cite{ranney1994models} were not accurate enough to be used in an autonomous vehicle's motion planning stack.
To remedy this, many researchers have been developing phenomenological methods, i.e., methods that learn the behavior of agents from a wealth of data (e.g., \cite{alahi2016social,kuefler2017imitating,MFP,desire}), to great effect.

In a typical autonomy stack, the trajectory prediction module is followed by a planning module which takes the predicted trajectories of surrounding agents and plans the ego-motion accordingly. With this downstream planner in mind, several requirements, in addition to prediction accuracy, emerge, and are discussed in detail below.

\begin{figure}
    \centering
    \includegraphics[width=1\columnwidth]{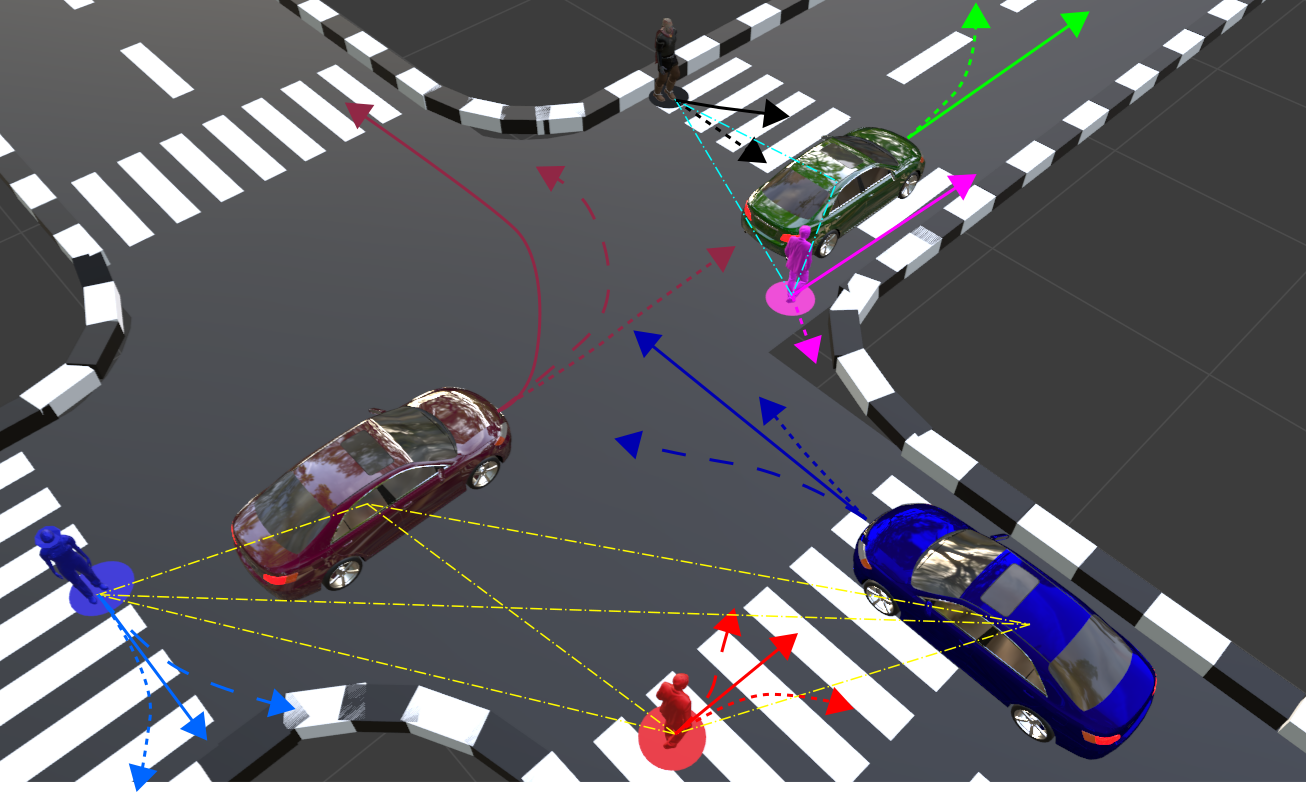}
    \caption{An illustration of \algname{}'s output, comprised of multimodal trajectory predictions for each agent. Different stroke type (solid, dashed, dotted) represents the different modes of the scene-consistent joint trajectory prediction. Agents in a scene are partitioned into highly-interacting cliques, an example of which is visualized with yellow dashed lines.}
    \label{fig:my_label}
    \vspace{-0.3cm}
\end{figure}

\subsection{Desiderata}\label{sec:desiderata}
Typical features desired for a trajectory prediction model
include high prediction accuracy, fast inference speed, and calibrated uncertainties. When predictions are subsequently consumed by a downstream planner, the following features are also critical for overall system performance:

\noindent\textbf{Compatibility:} Trajectory predictions for different agents in a scene should be compatible with each other within a single joint prediction. In particular, collisions among predicted trajectories should be rare, as collisions are themselves rare in reality. 

\noindent\textbf{Tractable Joint Trajectory Prediction:} 
As mentioned previously, the future motion of the agents can be multimodal. In a scene consisting of multiple agents, if the multimodal predictions are generated for individual nodes, a downstream motion planner needs to consider all combinations of these trajectory predictions. Since the number of modes grows exponentially with the number of agents, the planner is quickly overwhelmed. Alternatively, the motion planner could take a conservative approach and avoid all predicted trajectories, yet often at the price of compromised planning performance (e.g., bringing the robot to a standstill if all plans seem to collide). As a result, we desire multimodal, joint predictions of all agents with a limited, but fully-representative number of modes such that a downstream planner can perform contingency planning.

\noindent\textbf{Time Consistency:} With a downstream planner, the motion plan heavily depends on the prediction results. To ensure smooth motion plans, predictions should not change significantly between subsequent time steps if the scene itself has not changed drastically in the meantime. As a result, sampling should be avoided as predictions may change significantly between time steps, causing discontinuity in the resulting motion plans which may hurt planning performance and safety.

\noindent\textbf{Conditioning:} Conditioning is the operation of fixing one or multiple agents' future trajectories and predicting the resulting distribution of other agents' future trajectories. Conditional prediction is useful for motion planning (conditioned on the ego agent's motion plan) and for understanding interactions between agents. Conditioning is available in several existing works such as \cite{T++}, yet requires explicit modeling. Ideally, conditional distributions would be generated without requiring structural changes to the model.

\subsection{Contributions}
In this work we present \algname{}, a trajectory forecasting method that generates joint trajectory predictions for multiple interacting agents. Our contributions are threefold: First, we propose predicting the futures of \textit{cliques} of agents rather than individuals or the scene graph as a whole (\cref{sec:prepro}), and present a neural network architecture to do so (\cref{sec:encoder}).
Second, we leverage insights from motion planning and propose a policy network that autoregressively rolls out closed-loop trajectory predictions via a GNN that models agent-to-agent interactions and maps them to control inputs
(\cref{sec:decoder}).
Finally, we improve output sample diversity by augmenting our loss function with a tunable risk measure that determines weights between trajectory samples during training (\cref{sec:risk}).

When evaluated on large-scale, real-world pedestrian and driving datasets, \algname{} reduces the dimensionality necessary to capture scene-level multimodality (\cref{sec:peds}); achieves significant improvements in the scene consistency of its predictions, as measured by collision rate (\cref{sec:nusc}); and easily enables counterfactual analyses (\cref{sec:conditioning}); all of which are critical for simulation (\cref{sec:conditioning}), downstream planning (\cref{sec:planning}), and verification of autonomous vehicle performance.


\section{Related Work}
Early trajectory prediction works were predominantly ontological, positing structure about an agent's decision-making process, exemplified by social force models~\cite{helbing1995social,mehran2009abnormal}, hidden Markov models~\cite{kitani2012activity}, and the intelligent driver model~\cite{IDM}. However, limited by their expressibility, these models cannot scale to complicated scenarios despite extensive tuning.
To remedy this, many researchers have been developing phenomenological methods, i.e., methods that focus on learning the behavior of agents from a wealth of data. Several notable works include the Social LSTM \cite{alahi2016social}, GAIL \cite{kuefler2017imitating}, MFP \cite{MFP}, and DESIRE \cite{desire}.
Since the trajectory prediction problem is intrinsically a sequence-to-sequence modeling task, recent works commonly apply Recurrent Neural Networks (RNNs)~\cite{T++,ILVM,MFP,social_gan,sophie,desire} and Transformers~\cite{liu2021multimodal,ngiam2021scene,yuan2021agentformer}, achieving strong results.

Another core facet of trajectory prediction is accounting for the interactions between agents and the scene geometry. Two common choices to model agent interactions are Graph Neural Networks (GNNs) and Convolutional Neural Networks (CNNs). GNN-based methods~\cite{ILVM,T++,grouptron,ptp} construct scene graphs with agents as nodes and their interaction as edges, performing message passing to congregate information. CNN-based methods~\cite{MATF,predictionnet,ngiam2021scene,cui2019multimodal,djuric2020multixnet,chauffeurnet,phan2020covernet} typically rasterize scene information into layers of images such as Birds Eye View (BEV) images and velocity images to encode information. In general, CNN-based methods have a fixed computation complexity (usually being faster than GNN-based methods) and preserve geometric information about the scene. GNN-based methods can be viewed as a ``sparse" representation of the scene and allow for more complicated features, yet their computation complexity scales at least linearly with the number of agents, and geometric information may be lost without the use of special structures~\cite{spagnn}.

Once scene information is gathered and encoded, generative models such as GANs~\cite{social_gan,sophie} and CVAEs~\cite{CSP,T++,desire} are typically used to produce multimodal trajectory predictions.
Of these, CVAE models are the most common choice due to their performance and ease of training.
CVAEs with continuous latent spaces~\cite{ILVM,ptp,grouptron} enjoy stronger expressivity, but require sampling to obtain predictions, removing any time consistency in sequential outputs. On the other hand, a discrete latent space~\cite{MFP,T++} does not require sampling, but is less expressive and more likely to suffer from mode collapse.

Since trajectory prediction usually concerns multiple agents in a scene, the issue of scene consistency emerges, i.e., predictions of different agents should not collide with each other or with static obstacles. For models that pursue only high prediction accuracy, especially agent-centric models, scene consistency is usually poor. To remedy this, \cite{social_gan,deo2018convolutional} use pooling to model interaction between agents in the encoder, \cite{grouptron} introduces a group-level encoder with clustered agents sharing a group mode, and \cite{ILVM} learns a scene interaction model via message passing. Comparing to the encoder, coupled decoding is more difficult as the future trajectories are not known. \cite{ma2017forecasting} uses fictitious play to roll out predictions and gradually improve prediction quality, \cite{MFP} performs autoregressive decoding, building subsequent predictions on prior steps of prediction. \cite{spagnn} uses a message-passing procedure to search for most likely joint trajectories, yet only predicts a single mode. To the best of our knowledge, most existing methods do not model the agents as policy planners with observations, cost functions, and actuation inputs, all while explicitly enforcing scene consistency.

For autonomous vehicles which involve a downstream planner, it is also important to consider multimodality, i.e., the possibility for multiple distinct futures. 
However, the number of modes cannot be too large so as not to overwhelm the downstream planner. \cite{chen2020reactive} generates a set of motion primitives that are deemed possible and constrain the planner to avoid them. \cite{MATS} represents multimodal predictions as mixtures of linear dynamics (as opposed to tracklets), simplifying incorporation in downstream planning. To achieve a wide coverage of modes with a limited computational budget, several diversity sampling techniques have been proposed. \cite{DPP} proposes using determinantal point processes for diverse latent sampling, \cite{huang2020diversitygan} uses the farthest point sampling algorithm, and \cite{cui2021lookout} apply a reparameterization trick with coefficients given by a GNN. Overall, most existing diversity sampling techniques are designed for continuous latent spaces. In this work, we introduce a new risk-based loss modification that yields output diversity for a discrete latent space with only a small number of samples.

\section{\algname}
\algname{}\footnote{Code available at \url{https://github.com/nvr-avg/ScePT}} is a discrete CVAE model that outputs joint trajectory predictions for multiple agents in a scene, ensuring high scene consistency in its predictions by reasoning about each agent's motion policy and the influence of their neighbors.

\textbf{Nomenclature.}
Throughout the paper, we use the terms node and agent interchangeably, which may be a vehicle, a pedestrian, a cyclist, or other kinds of road users. We use $s$ to denote an agent's state and $e$ an edge between two nodes. Since our model is a CVAE, we follow standard terminology in the CVAE literature, i.e., $x$ denotes the conditioning variable, $y$ the observed variable, and $z$ the hidden latent variable. We use bold font to denote variables associated with a group of nodes, e.g., a clique. For example, for a clique consisting of nodes 1 through $N$, $z_1,...z_N$ are the latent variables of each of the nodes and $\mathbf{z}=[z_1,...z_N]$ is the latent variable of the clique.

\subsection{Preprocessing}\label{sec:prepro}
To maintain scene consistency, \algname{} is a scene-centric model, i.e., its output predictions are the joint trajectories of multiple nodes in a scene. Given a scene with multiple nodes, a spatiotemporal scene graph is generated where nodes represent agents and edges represent their interactions. We use agents' closest future distance as a proxy for interaction, 
propagating forward each node according to a constant velocity model $\Phi_{a_i=0}^{0:T}(s_i)$, where $\Phi_{a_i=0}^t$ is the flow operator that maps the initial state to the future states $t$ time steps ahead with the action $a_i$ of node $i$ set to 0 and $T$ is the prediction horizon. The closest future distance between two agents is defined as
\begin{equation}
    d_{ij} = \min_{t\in[0,T]} \text{Dis}(\Phi_{a_i=0}^t(s_i),\Phi_{a_i=0}^t(s_j)),
\end{equation}
where Dis is the Euclidean distance between the two agents. We then define the scene graph adjacency matrix as 
\begin{equation*}
    A_{ij}=\left\{\begin{array}{cc}
        0, & d_{ij}>d_0(\eta_i,\eta_j) \\
        \frac{d_0(\eta_i,\eta_j)}{d_{ij}} & d_{ij}\le d_0(\eta_i,\eta_j)
    \end{array} \right.,
\end{equation*}
where $\eta_i$ is the type of node $i$ (e.g., vehicle, cyclist, pedestrian) and $d_0$ is a distance threshold which is fixed for each edge type. 

With the scene graph determined by the adjacency matrix, in contrast to models that keep all nodes in a single graph~\cite{MFP,ILVM}, we partition the scene graph into cliques with a maximum size (fixed as a parameter). We do this to reduce the dimensionality of the product latent space, which scales exponentially with the size of the graph, causing prediction accuracy to deteriorate if too large (see \cref{sec:ablation} for further discussion). While weighted graph partitioning is NP-hard, there are many off-the-shelf algorithms, and we use the well-known Louvain algorithm due to its strong performance \cite{blondel2008fast}. After partitioning, every pair of nodes within a clique is connected (despite the distance threshold) to form a clique. Node histories
are then collected and fed to \algname. Node states and dynamics are explained in detail in \cref{sec:dyn}. When available, we also utilize the map information and the relative position to the closest lane.

\subsection{Encoder}\label{sec:encoder}

With cliques in hand, agents' state and edge (relative states between agents) histories are encoded into feature vectors via LSTMs. Instead of associating each node with a latent variable distribution that is independent of its neighbors, our encoder models the joint latent distribution. Specifically, each agent is equipped with a discrete latent variable $z_i$ with cardinality $N$, making the joint latent variable of the clique simply $\mathbf{z}=[z_1,z_2,...z_n]$. This means that the cardinality of the joint latent space grows exponentially with the number of nodes in the clique, and is the reason why we limit clique size. 

\begin{figure}
    \centering
    \includegraphics[width=0.4\columnwidth]{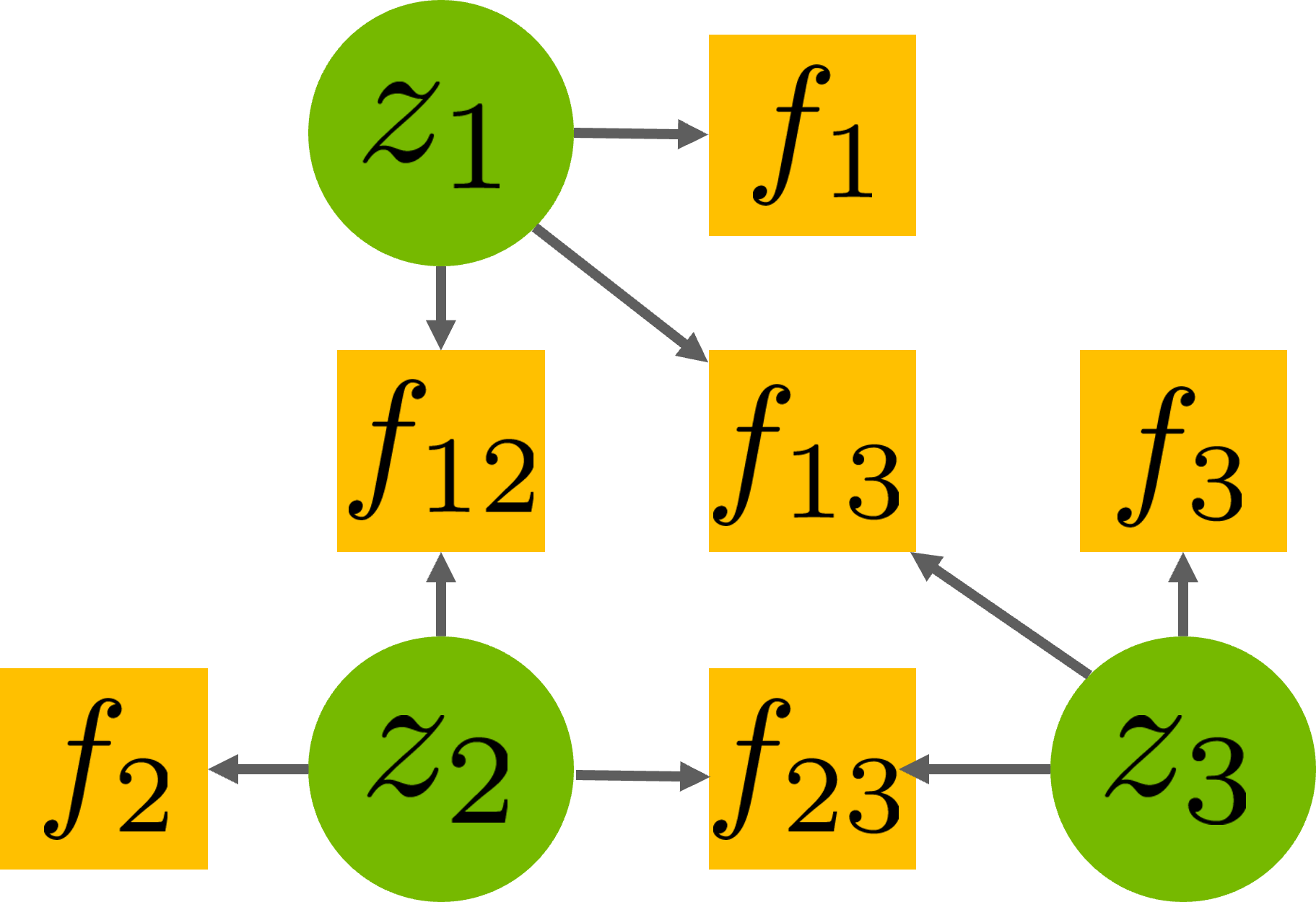}
    \caption{Factor graph with individual agent latent variables as variable nodes and factor nodes which are functions of the connected variable nodes. Factor nodes are comprised of individual agent and agent-agent interaction factors, e.g., $f_1$ is a function of $z_1$ while $f_{12}$ is a function of $z_1$ and $z_2$. All factor nodes are summed to obtain the log likelihood.}
    \label{fig:factor}
    \vspace{-0.3cm}
\end{figure}

\algname{} represents the distribution of the joint latent variable as a Gibbs distribution consisting of node factors and edge factors,
\begin{equation}
    \log \mathbb{P}(\mathbf{z}) \propto \sum_i f_i(x_i,z_i) + \sum_{e_{ij}\in\mathcal{E}} f_{ij}(x_i,x_j,z_i,z_j),
\end{equation}
where $x_i$ is the state history of node $i$ and $f_i$ is the node factor of node $i$, which is a feedforward neural network mapping $x_i$ and $z_i$ to a real number. $f_{ij}$ is the edge factor of the node pair $i, j$, also a feedforward network, and $\mathcal{E}$ is the set of edges. The log-likelihood can be computed by constructing a factor graph~\cite{abbeel2006learning}, which is a bipartite graph with variable nodes and factor nodes. An example factor graph is shown in \cref{fig:factor}. Normalization is done by summing up all possible valuations of $\mathbf{z}$ (since $\mathcal{Z}$ is discrete).

While the joint latent space's cardinality scales exponentially with clique size, we found that probability mass typically concentrates on only a few ($<10$) modes. 

\subsection{Decoder}\label{sec:decoder}

\begin{figure}
    \centering
    \includegraphics[width=1\columnwidth]{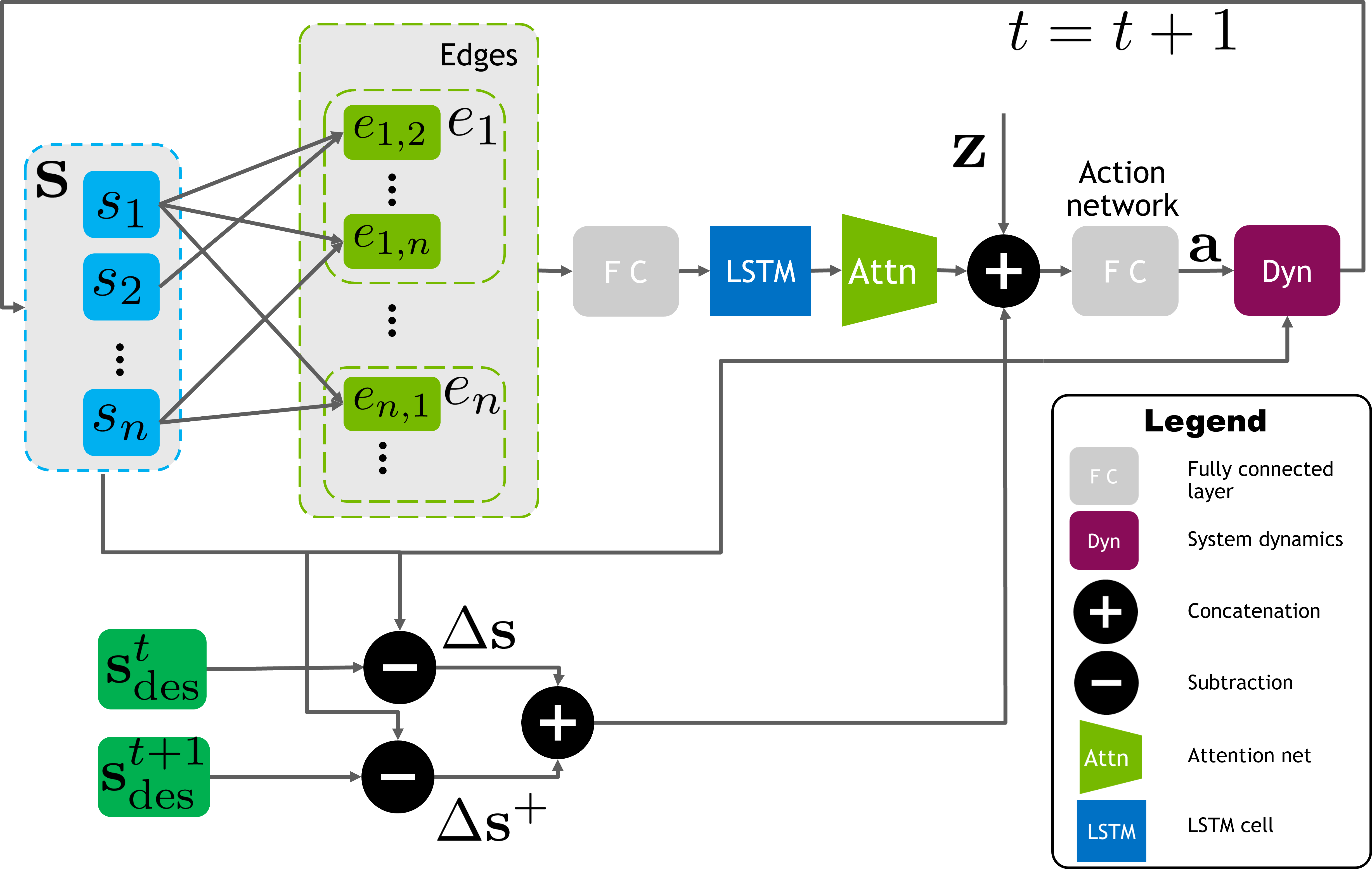}
    \caption{Our autoregressive policy network architecture. For each node, neighboring node states are pooled with an attention mechanism, the resulting encoding then generates control inputs together with the reference trajectory. The control inputs pass through the agent dynamics to produce position predictions and the process repeats for subsequent timesteps.}
    \label{fig:policy_net}
    \vspace{-0.4cm}
\end{figure}

Our decoder design is inspired by the motion planning process, i.e., we view each agent as a motion planner and emulate their planning process to output trajectory predictions. 
A typical motion planner takes a reference trajectory, i.e., the desired motion, and adjusts it to satisfy constraints (e.g., collision avoidance) and minimize a specified cost function. Inspired by this process, the structure of our policy net is visualized in \cref{fig:policy_net}.

The inputs to the policy net are the current states of the clique nodes, their reference trajectories $\mathbf{s}_{\text{des}}$, and the clique latent $\mathbf{z}$. Reference trajectories are generated via Gated Recurrent Unit (GRU) networks that take the state history encoding, map encoding, and latent variable $z$ as input. The current node states are then compared with the reference trajectories to obtain the tracking error $\Delta \mathbf{s}$ and the next waypoint in the local coordinate frame $\Delta \mathbf{s}^+$. 

To model an edge, we pair its two node states together and feed the state pair into a pre-encoding network (fully connected) and then an LSTM cell. For each node, depending on the graph structure, there may be a varying number of neighbors. To encode a variable number of neighbors, all of a node's edges are condensed into a single observation encoding via an attention network \cite{chorowski2015attention}. The observation encoding, latent variable, and tracking error are then concatenated and fed to a fully connected action network to obtain the node's control action prediction $a$. Here, we assume that the node's dynamics are differentiable functions of the state and control input,
which is true for common agent types such as vehicles (e.g., Dubin's car model~\cite{dubins1957curves}) and pedestrians (single or double integrators). The state prediction is then fed back to the state vector and this process repeats.

\begin{figure}
    \centering
    \includegraphics[width=1\columnwidth]{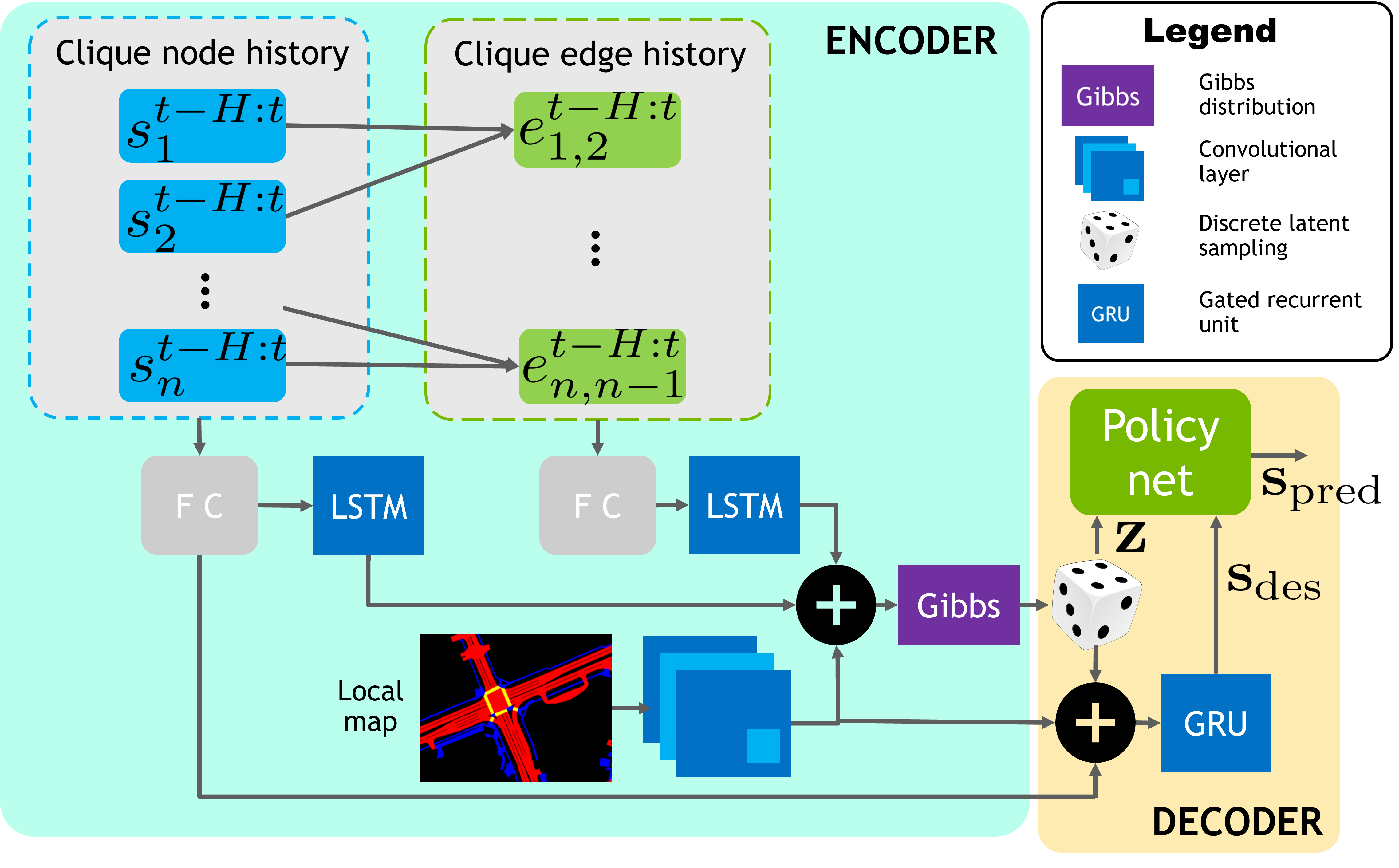}
    \caption{Overview of \algname: Node history and map information are collected for all nodes within a clique and passed through a Gibbs distribution to generate the discrete joint latent distribution. The policy net (decoder) then generates closed-loop trajectory predictions given latent samples.}
    \label{fig:overview}
    \vspace{-0.6cm}
\end{figure}

The overall structure of \algname{} is shown in \cref{fig:overview}. The encoder takes the LSTM-encoded state and edge history as well as the CNN-encoded local map, and generates a discrete Gibbs distribution over the clique latent variable. The latent variable, together with the state history and map encodings, is used to generate the desired trajectory for each node via GRUs. The desired trajectories and latent variable are then passed to the policy net to obtain closed-loop trajectory predictions.

\subsection{Conditioning via policy learning}
As mentioned in \cref{sec:desiderata}, conditioned prediction is an important capability. Conditioning was performed in previous works~\cite{ivanovic2019trajectron,T++} by explicitly encoding the ego future trajectory in the encoder. However, assuming that only one agent can be conditioned on makes use cases such as driving simulation difficult, as one would need to train explicit conditioning models for each pair of agents. By comparison, PRECOG~\cite{rhinehart2019precog} simply needs to set the latent variable of the robot to produce future-conditional predictions. Similarly, \algname{} does not require any structural change to produce conditional predictions since it learns agents' interaction policies. Conditional predictions are generated by simply fixing the trajectory roll-outs of conditioned agents and outputting the trajectory predictions of the rest of the agents in the clique. Since a fixed future trajectory does not fall into any latent mode, we remove any factors concerning conditioned nodes from the Gibbs distribution factor graph.

\subsection{Training}
Following standard CVAE training~\cite{sohn2015learning}, our objective is the Evidence Lower Bound (ELBO) loss:
\begin{equation}
\begin{aligned}
    ELBO &= \mathbb{E}_{\mathbf{z}\sim Q(\mathbf{z}|\mathbf{x},\mathbf{y})} [\log(P(\mathbf{y}|\mathbf{x},\mathbf{z}))] \\& - \beta D_{KL}(Q(\mathbf{z}|\mathbf{x},\mathbf{y})||P(\mathbf{z}|\mathbf{x})),
    \end{aligned}
\end{equation}
where $\mathbf{z}$ is the clique latent variable, $\mathbf{y}$ is the future trajectories of all nodes, and $\mathbf{x}$ is the conditional variable, consisting of node and edge history, map encoding, and lane information for all nodes in the clique. 
For the likelihood cost, we assume Gaussian noise around the predicted trajectory for each mode, resulting in a 2-norm loss,
\begin{equation}\label{eq:expectation_loss}
\begin{aligned}
    &\mathbb{E}_{\mathbf{z}\sim Q(\mathbf{z}|\mathbf{x},\mathbf{y})} [\log(P(\mathbf{y}|\mathbf{x},\mathbf{z}))] \\
    =& \sum_{\mathbf{z}\in\mathcal{Z}} Q(\mathbf{z}|\mathbf{x},\mathbf{y}) ||f_{\mathbf{y}}(\mathbf{x},\mathbf{z})-\mathbf{y}_{GT}(\mathbf{x})||^2,
    \end{aligned}
\end{equation}
where $f_{\mathbf{y}}(\mathbf{x},\mathbf{z})$ is the trajectory prediction from the decoder and $\mathbf{y}_{GT}(\mathbf{x})$ is the ground truth. 

We also add a collision penalty, specified in detail in \cref{sec:collision}, as a regularization term to penalize incompatible predictions, the influence of which will be further discussed in \cref{sec:ablation}. Other types of regularization, e.g., ride comfort, can also be added since the node dynamics are explicitly included in the policy net.

\noindent\textbf{Sampling the Latent Space.} While our discrete latent space is enumerable, the cardinality of $\mathcal{Z}$ grows exponentially with the clique size. Thus, it is sometimes not tractable to decode all modes. To remedy this, we apply diversity sampling. Specifically, we take the $N_{g}$ highest probability
modes and randomly sample $N_r$ 
modes from the rest. When the total cardinality of $\mathcal{Z}$ is less than $N_g+N_r$, all modes are selected. The sample probabilities are then normalized
so that the expected loss does not collapse to 0.

\subsection{Mode Collapse and Diverse Sampling}\label{sec:risk}
Discrete CVAEs for trajectory prediction are prone to mode collapse, i.e., the decoder tends to predict similar trajectories under different modes since the likelihood cost is a weighted sum of 2-norm errors and the average prediction is likely to be a local minimum. Mode collapse has been discussed in previous works and tackled by methods such as Multiple-Trajectory Prediction (MTP) loss~\cite{cui2019multimodal}, using prior knowledge~\cite{djuric2020multixnet,casas2020importance}, and assigning modes by classifying the ground truth into categories ~\cite{liu2021multimodal}. Our approach maintains the expected loss function, but introduces CVaR as a new way to avoid mode collapse.

\noindent\textbf{Conditional Value at Risk} (CVaR) \cite{rockafellar2000optimization} is a risk measure commonly used in finance and optimization, defined as
\begin{equation}\label{eq:CVaR}
\begin{aligned}
  \text{CVaR}_{1-\alpha}(X)&=\inf_{\eta\in\mathbb{R}}\{\eta+\frac{1}{\alpha}\int_{-\infty}^\infty [x-\eta]_+P(x)\} \\
  &=\min_{0\le P'(x)\le \frac{1}{\alpha}P(x),\int P'(x)dx=1}\mathbb{E}_{P'}[X],
  \end{aligned}
\end{equation}
where $P$ is the probability distribution of $X$ and $\alpha$ tunes the level of risk-averseness. CVaR is the mean of the lowest $\alpha$-percentile values of $x$ under $P$. At the limits of $\alpha$, 
$\alpha \rightarrow 0$ yields the essential infimum of $X$ and $\alpha = 1$
yields $\mathbb{E}[X]$.

The second line in \eqref{eq:CVaR} is the dual form of CVaR, which can be understood as shifting the distribution $P$ to $P'$ under the constraint that $P'$ has to be a proper distribution and for all $x$, $P'(x)\le \frac{1}{\alpha} P(x)$. Inspired by the dual form, we modify the expectation loss in \eqref{eq:expectation_loss} to
\begin{equation}
    \min_{0\le Q'(z)\le\frac{1}{\alpha} Q(z|x,y),\sum Q'(z)=1} \mathbb{E}_{z\sim Q'} [||f_y(x,z)-y_{GT}(x)||^2],
\end{equation}
which is the best $\alpha$-percentile loss value among the discrete modes. This CVaR loss does not force all modes to match the ground truth, only those that are already close, directly preventing mode collapse. Compared to common usages of risk measures which typically focus on the worst outcomes, we use CVaR to focus on the best predictions to maintain output diversity. During training, $\alpha$ is used to trade-off the model's focus on encoder accuracy vs diversity, see \cref{sec:alpha} for details.
In addition to incorporating CVaR, we also use a greedy algorithm to diversely sample the product latent space, see \cref{sec:product_sampling} for further details.


\section{Experiments}
We evaluate the performance of \algname{} on the tasks of pedestrian and vehicle motion prediction. In particular, we make use of the well-known ETH \cite{pellegrini2009you}, UCY \cite{lerner2007crowds},
and nuScenes \cite{caesar2020nuscenes} datasets. 

\noindent \textbf{Metrics.} We use the common Average and Final Displacement Error (ADE/FDE) metrics to measure the quality of trajectory predictions. Since our outputs are multimodal, we adopt the Best-of-N (BoN) extension and take the $N$ highest probability modes from the encoder to compute BoN ADE/FDE values. The sampling process in \algname{} is different from prior works with continuous latent spaces as we do not randomly sample from the latent distribution, but rather pick the $N$ modes deemed most likely by the encoder. As a result, when the number of samples is larger than the number of possible clique modes (i.e., 
$N>|\mathcal{Z}|$), we take only $|\mathcal{Z}|$ samples.

\subsection{Pedestrian Motion Prediction}\label{sec:peds}
The ETH (containing ETH and Hotel scenes) \cite{pellegrini2009you} and UCY (containing Univ, Zara1 and Zara2 scenes) \cite{lerner2007crowds} datasets are widely-used benchmarks for pedestrian motion prediction. Together, they contain 9,514 unique pedestrians in many challenging, interactive real-world scenarios.

For all pedestrian datasets, the maximum clique size is 5 and each node's latent space cardinality is 6. The ADE/FDE results are shown in Tables \ref{tab:ADE1} (deterministic) and \ref{tab:ADE2} (multimodal).
While \algname{} is designed mainly for autonomous driving, it performs remarkably well on pedestrian datasets, achieving the best or second-best performance among state-of-the-art models in the field. In particular, \algname{} outperforms prior methods on most datasets in ADE, yet performs slightly worse in UCY scenes in terms of FDE. The likely reason is that the UCY datasets are much denser than ETH, forcing \algname{} to partition the large scene graph into small cliques. Once partitioned, interactions between cliques are ignored, hurting prediction accuracy. Using a larger maximum clique size, however, would cause the joint latent cardinality to be too large and further deteriorate performance. Even in these cases, \algname{} still performs on par with state-of-the-art prediction methods.

\begin{table*}[]
\centering
\begin{tabular}{c|ccc|c}
Dataset & S-LSTM~\cite{alahi2016social}    & S-ATTN~\cite{VemulaMuellingEtAl2018}    & Trajectron++~\cite{T++} & \algname{} \\ \hline
ETH     & 1.09/2.35 & \textit{0.39}/3.74 & 0.71/\textit{1.66}    & \textbf{0.19}/\textbf{1.33}         \\
Hotel   & 0.79/1.76 & 0.29/2.64 & \textit{0.22}/\textbf{0.46}    & \textbf{0.18}/\textit{1.12}         \\
Univ    & 0.67/1.40 & \textit{0.33}/3.92 & 0.44/\textbf{1.17}    & \textbf{0.19}/\textit{1.19}         \\
Zara1   & 0.47/1.00 & \textit{0.20}/\textbf{0.52} & 0.30/\textit{0.79}     & \textbf{0.18}/1.10          \\
Zara2   & 0.56/\textit{1.17} & 0.30/2.13 & 0.23/\textbf{0.59}    & \textbf{0.19}/1.20 \\
\hline
Average & 0.71/1.54 &	0.30/2.59 &	\textit{0.38}/\textbf{0.93} &	\textbf{0.19}/\textit{1.19}
\end{tabular}
\caption{ADE/FDE in meters, using the most likely mode.  \textbf{Bold}/\textit{italic} font indicates the best/second-best value. Lower is better.}\label{tab:ADE1}
\vspace{-0.2cm}
\end{table*}
\begin{table*}[]
\centering
\begin{tabular}{c|cccc|c}
Dataset & S-GAN~\cite{social_gan}     & SoPhie~\cite{sophie}    & MATF~\cite{MATF}      & Trajectron++~\cite{T++} & \algname{} \\ \hline
ETH     & 0.81/1.52 & 0.70/1.43 & 1.01/1.75 & \textit{0.39}/\textit{0.83}    & \textbf{0.10/0.65}         \\
Hotel   & 0.72/1.61 & 0.76/1.67 & 0.43/0.80 & \textbf{0.12/0.21}    & \textit{0.13}/\textit{0.77}         \\
Univ    & 0.60/1.26 & 0.54/1.24 & 0.44/0.91 & \textit{0.20}/\textbf{0.44}    & \textbf{0.12}/\textit{0.65}         \\
Zara1   & 0.34/0.69 & 0.30/0.63 & 0.26/\textit{0.45} & \textit{0.15}/\textbf{0.33}    & \textbf{0.13}/0.77         \\
Zara2   & 0.42/0.84 & 0.38/0.78 & 0.26/\textit{0.57} & \textbf{0.11/0.25}    & \textit{0.14}/0.81 \\
\hline
Average& 0.58/1.18&	0.54/1.15&	0.48/0.90&	\textit{0.19}/\textbf{0.41}&	\textbf{0.12}/\textit{0.73}
\end{tabular}
\caption{Best-of-20 ADE/FDE in meters. \textbf{Bold}/\textit{italic} font indicates the best/second-best value. Lower is better.}\label{tab:ADE2}
\vspace{-0.2cm}
\end{table*}

\begin{figure}
    \centering
    \includegraphics[width=1\columnwidth]{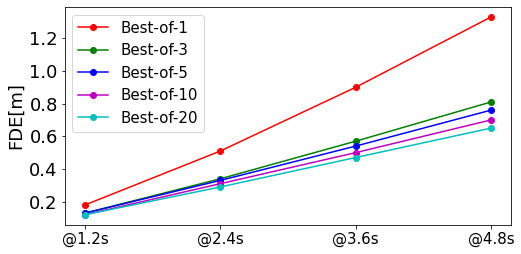}
    \caption{FDE of the ETH dataset under varying numbers of samples, the most significant improvement is seen at 3 samples.}
    \label{fig:BO}
    \vspace{-0.5cm}
\end{figure}
\cref{fig:BO} shows the FDE of the ETH dataset under different numbers of samples, Due to our use of CVaR in the loss function, \algname{} is able to generate diverse modes. After only 3 samples, our method's predictions are already very accurate.
We observe this phenomenon across all datasets, and find that sampling 3 to 5 modes achieves a good balance of prediction quality and runtime complexity.

\subsection{Vehicle Motion Prediction}\label{sec:nusc}
The nuScenes dataset \cite{nuscenes} consists of 1000 driving scenes, each 20 seconds long and containing up to 23 object classes. To match the nuScenes prediction challenge set, only vehicles and pedestrians are predicted during training and evaluation. \cref{tab:nusc} summarizes the prediction accuracy of our method alongside a set of state-of-the-art approaches. Despite \algname{} discarding edges between cliques, the prediction accuracy is near the state-of-the-art, especially in later timesteps. Furthermore, thanks to our diversity-promoting design, prediction accuracy can be significantly improved by adding only 1 or 2 extra modes.

\begin{table}[]
\centering
\begin{tabular}{c|cccc}
Method              & @1s  & @2s  & @3s  & @4s  \\ \hline
S-LSTM\cite{alahi2016social,spagnn}   & 0.47 &    -  & 1.61 &  -    \\
CSP\cite{CSP,spagnn}              & 0.46  & -     & 1.50 & -     \\
CAR-Net\cite{Carnet,spagnn}          & 0.38 & -     & 1.35 & -     \\
SpAGNN\cite{spagnn}           & 0.35 & -     & 1.23 & -    \\
Trajectron++\cite{T++}     & 0.07 & 0.45 & 1.14 & 2.20  \\
\hline
Ours             & 0.44 & 0.93 & 1.63  & 2.58  \\
Ours (Best-of-2) & 0.41 & 0.83 & 1.44 & 2.20  \\
Ours (Best-of-3) & 0.40  & 0.80  & 1.36 & 2.14
\end{tabular}
\caption{\algname{}'s prediction accuracy on nuScenes vehicles nearly matches state-of-the-art methods. Furthermore, thanks to its diversity-promoting design, \algname{}'s prediction accuracy can be drastically improved by adding only 1 or 2 extra modes.}\label{tab:nusc}
\vspace{-0.5cm}
\end{table}

\begin{figure}
    \centering
    \includegraphics[width=1\columnwidth]{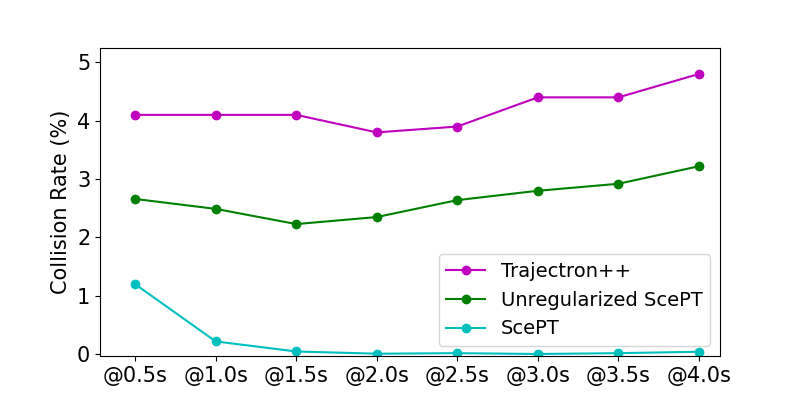}
    \caption{\algname{} achieves a better collision rate than Trajectron++~\cite{T++} without collision cost regularization. With collision cost regularization, the collision rate becomes virtually zero.}
    \label{fig:collision_rate}
    
    \vspace*{-0.4cm}
    
\end{figure}

As mentioned in the introduction, scene consistency in trajectory predictions is critical for planning, even more so when simulating and verifying the performance of autonomous vehicles. \cref{fig:collision_rate} compares the collision rate achieved by Trajectron++~\cite{T++} and \algname{} at different prediction horizons. The collision rate is averaged over all vehicles whose trajectories are predicted by the two models, including vehicle-to-vehicle and vehicle-to-pedestrian collisions. The result shows that \algname{}'s predictions have much less collisions than Trajectron++, even without collision cost regularization; becoming virtually zero with regularization. Moreover, the collision rate tends to decrease with the horizon, implying that collisions at earlier timesteps may be due to poor initial conditions and that the learned policy net is able to resolve conflicts in later timesteps.

\subsection{Conditioning and Counterfactual Analyses}\label{sec:conditioning}

\begin{figure}
    \centering
    \includegraphics[width=1\columnwidth]{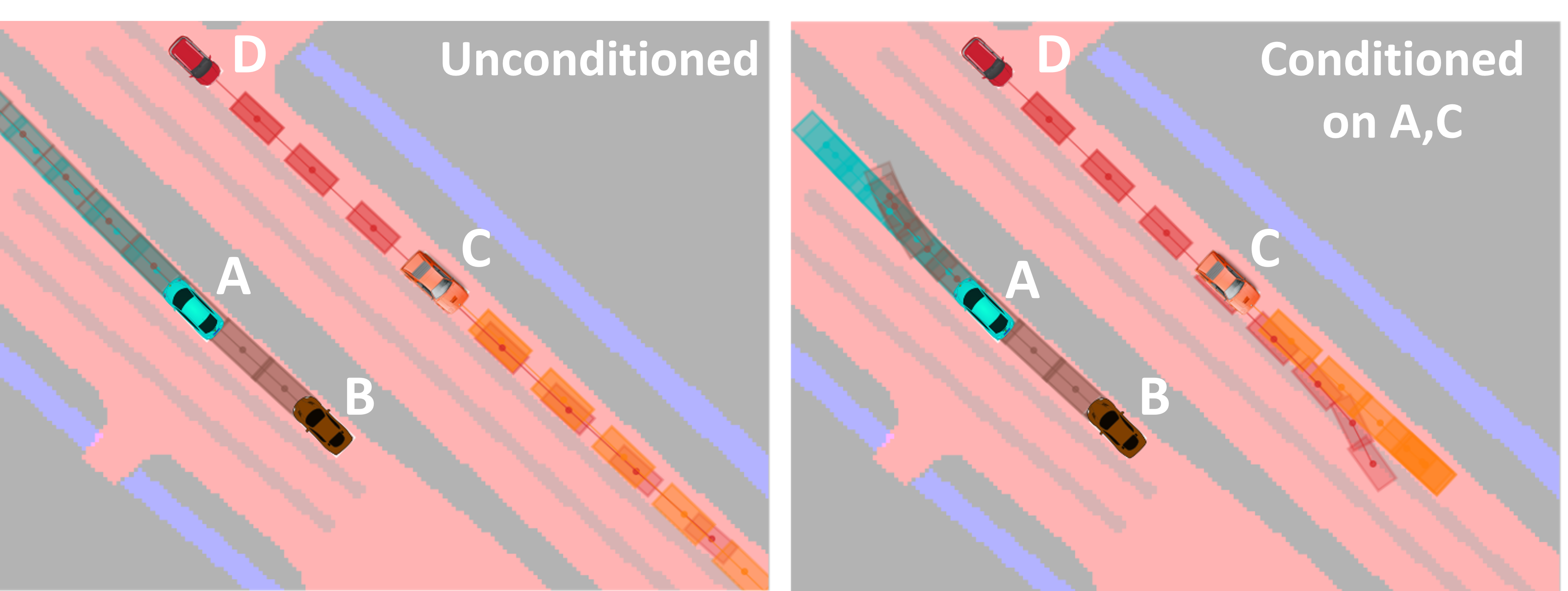}
    \includegraphics[width=1\columnwidth]{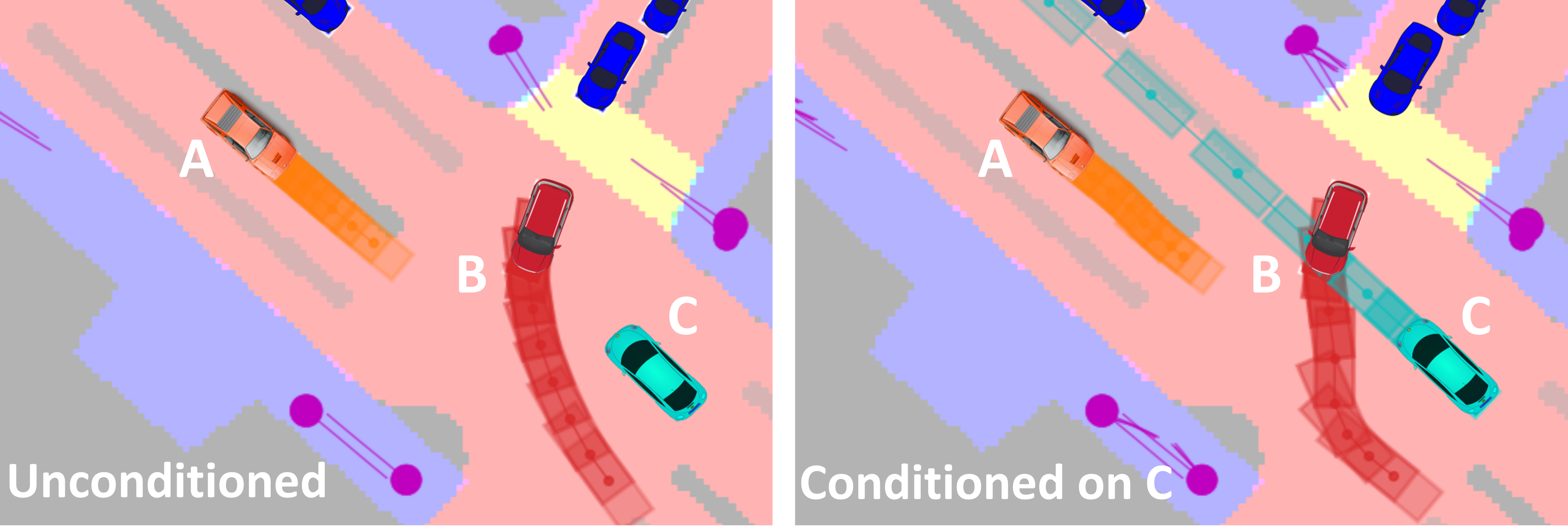}
    \caption{By design, \algname{} can produce predictions that are conditioned on any number of agents. \textbf{Left:} Original, unconditioned predictions. \textbf{Top Right:} Predictions conditioned on vehicles A and C braking. \textbf{Bottom Right:} Predictions conditioned on vehicle C accelerating.}
    \label{fig:cond}
    \vspace{-0.6cm}
\end{figure}
Conditioning is an important capability,  enabling a downstream planner to obtain trajectory predictions conditioned on ego motion. It is also useful for performing counterfactual, ``what if?" analyses. \cref{fig:cond} demonstrates \algname{}'s ability to perform conditional prediction, where the left and right figures show unconditioned and conditioned predictions, respectively. In the top right, we condition on vehicle A and C braking at $-4m/s^2$ to a full stop. Accordingly, our method predicts that (1) vehicle B will brake to avoid a collision, and (2) vehicle D will perform a lane change to avoid vehicle C. In the bottom right, we condition on vehicle C accelerating at $4m/s^2$. This causes a chain reaction, making vehicle B's prediction swerve to avoid a collision, which then subsequently affects vehicle A and the two pedestrians on the bottom left, showing that nodes within a clique are influenced by all other nodes in that clique.

Interestingly, the policy learned from the nuScenes dataset is quite robust to clique size. Our model is currently trained with a maximum clique size of 4, but we evaluated it with sizes as large as 8. Even in such cases, the model is still able to produce reasonable conditioned predictions, verifying the efficacy of the policy net's attention network.

\subsection{Integration with a Downstream Planner}\label{sec:planning}
\begin{figure}[t]
\includegraphics[width=1\linewidth]{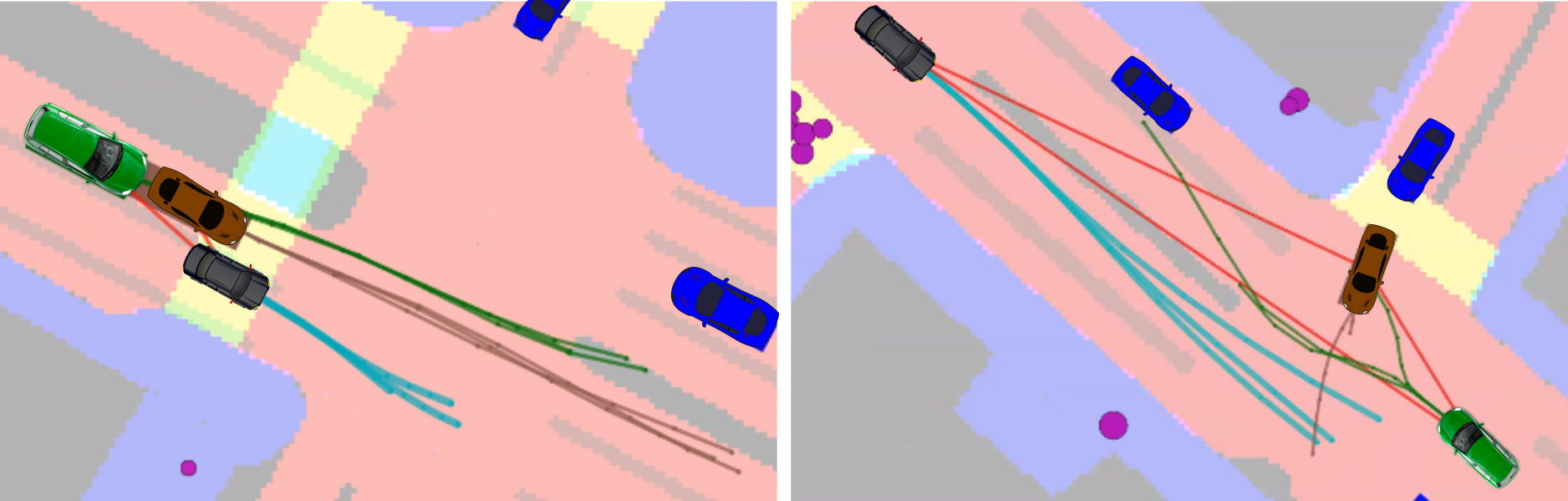}  
\caption{The integration of prediction and planning. \textbf{Black vehicle}: ego vehicle; \textcolor{blue}{blue vehicles}: adjacent vehicles outside the ego clique; \textcolor{cyan}{cyan trajectory}: planned trajectory (3 modes); \textcolor{green}{green} and \textcolor{brown}{brown} vehicles: adjacent vehicles within the ego clique; \textcolor{green}{green} and \textcolor{brown}{brown} trajectories: predicted trajectories (top 3 modes); \textcolor{magenta}{magenta circle}: pedestrians; \textcolor{red}{red lines}: connected nodes within the ego's clique.}
\label{fig:MPC}
\end{figure}

To demonstrate \algname's performance when integrated with a downstream planner, we feed its predictions to a downstream model predictive control (MPC)-based planner. The MPC planner takes the multimodal trajectory predictions into account and performs contingency planning via branching~\cite{alsterda2019contingency,chen2021interactive}. Given $M$ joint trajectory predictions, the MPC plans $M$ corresponding ego trajectories with the additional constraint that the first control inputs for all $M$ ego trajectories must be the same. Formally,
\begin{equation}\label{eq:mpc}
\begin{aligned}
\min_{\{a^i_{0:T-1},s^i_{0:T}\}_{i=1}^M} & \sum_i \pi_i\mathcal{J}(a^i_{0:T-1},s^i_{1:T})\\
\mathrm{s.t.}~& s^i_{t+1}=\text{Dyn}(s^i_t,a^i_t),\;\forall i, t, a^i_t\in \mathcal{A}, \; x^i_t\in\mathcal{S},\\
&  \mathcal{C}(s^i_{1:T},y^i_{1:T})\le 0, i=1...,M,\\ &s^i_0= s_0,\; a^1_0=a^2_0=...a^M_0,
\end{aligned}
\end{equation}
where $\pi_i$ is the probability of prediction mode $i$, $s^i$ and $u^i$ are the planned state and input sequences of the ego vehicle under the $i$-th mode, $\mathcal{J}$ is the cost function, and $\mathcal{C}$ is a constraint (e.g. collision avoidance).
\cref{eq:mpc} is a nonlinear optimization problem and is solved with IPOPT \cite{ipopt}. As an example of runtime, when $M=3$, our unoptimized PyTorch prediction code executes in less than 240ms and MPC planning takes less than 60 ms, all on a CPU. \cref{fig:MPC} shows the result of combining \algname{}'s predictions with this downstream MPC planner, visualizing prediction modes and their resulting ego motion plans.

\subsection{Ablation study}\label{sec:ablation}
\noindent\textbf{Collision Cost Regularization and Accuracy.}
As we have seen in \cref{sec:nusc}, the collision rate of \algname{} without the collision penalty already outperforms prior works. When collision penalty is added, the collision rate of \algname{} drops to virtually zero.
\begin{table}[]
\centering
\begin{tabular}{c|cccc}
                   & @1s  & @2s  & @3s & @4s  \\ \hline
With collision penalty    & 0.44 & 0.93 & 1.63 & 2.58  \\
Without collision penalty & 0.40  & 0.92 & 1.70 & 2.71
\end{tabular}
\caption{Influence of including collision penalty regularization on nuScenes vehicle prediction accuracy.}\label{tab:collision_ablation}
\end{table}
\cref{tab:collision_ablation} summarizes the effect of including the collision penalty on prediction accuracy. 
We can see that prediction accuracies are either the same or better after the collision penalty is added, indicating that avoiding collisions also produces more accurate outputs.

We also assess conditional predictions without the collision penalty, and collisions are much more likely when some nodes are assigned errant behavior. For example, the situation in \cref{fig:cond} (top) results in vehicles B and D hitting their lead vehicles when predicted by a model trained without collision penalty regularization. This phenomenon implies that, without collision penalty regularization, \algname{} is not able to maintain scene consistency in out-of-distribution scenarios. Overall, this regularization term forces the model to maintain scene consistency, making it more robust to out-of-distribution scenarios.

\noindent\textbf{Effect of $\alpha$.} To study the utility of using CVaR in \cref{eq:expectation_loss}, we conduct an ablation study over different values of $\alpha$. Our baseline model uses a varying $\alpha$ (initialized at 0.2 and gradually increasing to 1.0 during training). We also train a version with constant $\alpha = 1.0$, in which case the CVaR loss function is equivalent to the original expectation. \cref{tab:alpha_ablation} summarizes the results, showing that using CVaR outperforms expectation in the original loss function.
\begin{table}[]
\centering
\begin{tabular}{c|cccc}
  Loss Ablation       & @1s   & @2s  & @3s & @4s  \\ \hline
With CVaR     & 0.44  & 0.93 & 1.63 & 2.58  \\
Without CVaR & 0.45 & 0.97 & 1.74 & 2.79
\end{tabular}
\caption{Influence of CVaR loss on nuScenes prediction FDE.}\label{tab:alpha_ablation}
\vspace{-0.5cm}
\end{table}

\noindent\textbf{Effect of Clique Formation.}
To assess the sensitivity of \algname{} to the clique forming process, we change the distance criteria described in \cref{sec:prepro} to the Euclidean distance between nodes at the current time. \cref{tab:dis_ablation} summarizes the results, which shows that the flow distance partition leads to better performance, yet using a simple Euclidean distance leads to decent performance.

\begin{table}[]
\centering
\begin{tabular}{c|cccc}
     Distance Ablation          & @1s   & @2s  & @3s & @4s  \\ \hline
Flow   & 0.44  & 0.93 & 1.63 & 2.58  \\
$t=0$ Euclidean & 0.43 & 0.95 & 1.70 & 2.71
\end{tabular}
\caption{Influence of distance criteria for clique forming on nuScenes prediction accuracy (FDE).}\label{tab:dis_ablation}
\end{table}

Another important parameter is the maximum clique size. Since a node only considers neighbors within its clique, a small clique size leads to the nodes overlooking some nearby neighbors that were not partitioned into the same clique. On the other hand, if a clique gets too big, the cardinality of the clique's product latent space becomes too large, causing problems when sampling greedily. After experimenting, we found that a maximum clique size of 4 leads to the best result. \cref{tab:clique_size_ablation} summarizes this ablation.

\begin{table}[]
\centering
\begin{tabular}{c|cccc}
  Clique Size Ablation     & @1s   & @2s  & @3s & @4s  \\ \hline
Max clique size 2 & 0.41 & 0.90 & 1.63 & 2.68 \\
Max clique size 4    & 0.44  & 0.93 & 1.63 & 2.58  \\
Max clique size 6 & 0.64 & 1.27 & 1.90 & 2.90
\end{tabular}
\caption{Effect of maximum clique size on nuScenes prediction accuracy (FDE).}\label{tab:clique_size_ablation}
\vspace{-0.5cm}
\end{table}

\section{Conclusion and Future Work}\label{sec:conclusion}

\noindent{\bf Summary.}
This paper presents \algname{}, a CVAE-based model that generates multimodal joint trajectory predictions with high scene consistency. The encoder uses a Gibbs distribution to capture interactions between agents and outputs prediction mode probabilities for a whole clique instead of individual nodes. The decoder learns agent interaction policies to generate closed-loop trajectory predictions with high scene consistency, thanks to an explicit collision penalty used as regularization during training. Experiments on the ETH, UCY, and nuScenes datasets show that \algname{} is able to achieve state-of-the-art prediction accuracy with significantly improved scene consistency. We also demonstrate its capability to integrate with a downstream contingency planner and to generate human-like behaviors via conditioning.

\noindent{\bf Limitations and Future Work.} 
One important limitation (and area of future work) of \algname{} is its loss of sparsity. Since the decoder generates predictions concurrently for all nodes in a clique, one cannot utilize the sparsity of the interaction graph to only consider neighbors for each node. This causes the exponential cardinality of the joint latent space and forces the size of cliques to be limited. Subsequently, interactions between nodes in different cliques are ignored, leading to compromises in accuracy especially when there are large numbers of crowded agents, such as in the UCY dataset. Another limitation is the computation time due to the autoregressive policy net, which could be further improved by more efficient coding and parallelization.


{\small
\bibliographystyle{ieee_fullname.bst}
\bibliography{mybib}
}

\clearpage
\newpage
\appendix
\section{Preprocessing and Dynamics}
\subsection{Pre-encoding of Nodes and Edges}
Since all datasets we use provide global coordinates and headings of the nodes, we first transform the global coordinates of the state history and state future to local frames fixed at the nodes' current position. For vehicles with heading angle $\psi$, we use $\cos(\psi)$ and $\sin(\psi)$ as features instead of $\psi$ to prevent the $\pm 2\pi$ issue. For every type of nodes, the raw features are passed through a fully connected layer as pre-encoding, and the network is shared in all modules that require state information, i.e., whenever we use state information, the state vector passes through the pre-encoding layer first.

For edges, we construct a pre-encoding layer for every edge type (e.g., vehicle-vehicle, vehicle-pedestrian, pedestrian-vehicle, and pedestrian-pedestrian). The pre-encoding layer extracts raw features from the states of the two agents then passes them through a fully connected layer. The raw features contain agents' relative position and relative velocity in the local frame as well as their sizes.

\subsection{Dynamic Models and Collision Checking}\label{sec:dyn}
We use a Dubin's car model for vehicles in the scene with 

$s=\begin{bmatrix}X\\Y\\v\\\psi
\end{bmatrix}, a= \begin{bmatrix}\dot{v}\\ \dot{\psi}\end{bmatrix}, s^+=\begin{bmatrix}X+v\cos(\psi)\Delta t\\Y+v\sin(\psi)\Delta t\\v+\dot{v} \Delta t\\\psi+\dot{\psi}\Delta t\end{bmatrix}$
where $v$ and $\dot{v}$ are the longitudinal velocity and acceleration, $\psi$ and $\dot{\psi}$ are the heading angle and yaw rate.

The pedestrians follow a double integrator model with 

$s=\begin{bmatrix}X\\Y\\v_x\\v_y
\end{bmatrix}, a= \begin{bmatrix}\dot{v}_x\\ \dot{v}_y\end{bmatrix}, s^+=\begin{bmatrix}X+v_x\Delta t\\Y+v_y\Delta t\\ v_x+\dot{v}_x \Delta t\\v_y+\dot{v}_y\Delta t\end{bmatrix}$ .

Indeed, both models consists of basic differentiable functions that can be incorporated in a neural network. We also put bound on the inputs $\dot{v}\in[-5m/s,5m/s],\dot{\psi}\in[-1m/s^2,1m/s^2]$, $v_x,v_y\in[-5m/s,5m/s]$ so that the generated trajectory predictions are dynamically feasible.

\subsection{Collision Check}\label{sec:collision}
\begin{figure}
    \centering
    \includegraphics[width=1\columnwidth]{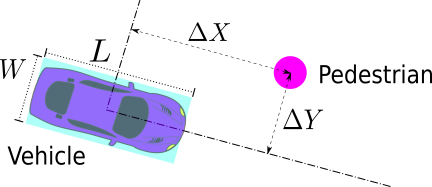}
    \caption{Collision check between a vehicle and a pedestrian}
    \label{fig:vp_col}
\end{figure}

We model pedestrians as circles with varying radius and vehicles as rectangles. The collision between pedestrians is straightforward to check, simply by taking the Euclidean distance between the two pedestrians. Collisions involving vehicles are checked in the local coordinate frame of the vehicle. \cref{fig:vp_col} shows the case with a pedestrian and a vehicle, and the collision function is 
\begin{equation*}
    \text{Col}(\Delta X,\Delta Y,L,W) = \max\{|\Delta X|-\frac{L}{2},|\Delta Y|-\frac{W}{2}\}.
\end{equation*}

\begin{figure}
    \centering
    \includegraphics[width=1\columnwidth]{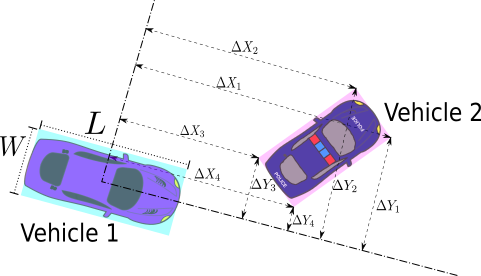}
    \caption{Collision check between two vehicles}
    \label{fig:vv_col}
\end{figure}

Vehicle-to-vehicle collision is a bit tricky since it involves two rectangles. As shown in \cref{fig:vv_col}, we use the four corners to calculate the $\text{Col}$ function:
\begin{equation*}
\begin{aligned}
    &\text{Col}(\Delta X_{1:4},\Delta Y_{1:4},L,W) \\=&\max\left\{\begin{aligned}|&\Delta X_1|-\frac{L}{2},...|\Delta X_4|-\frac{L}{2},\\&|\Delta Y_1|-\frac{W}{2},...|\Delta Y_4|-\frac{L}{2}\end{aligned}\right\}.
\end{aligned}
\end{equation*}
Note that the collision functions are all differentiable (at least piecewise differentiable), making it convenient to include them in the training process as regularization.

\subsection{Diversity Scheduling during Training}\label{sec:alpha}
The parameter $\alpha$ serves as a tuning knob to adjust the tradeoff between encoder accuracy and diversity. During training, we start with a small $\alpha$ so that the decoder can learn diverse trajectory patterns without mode collapse, then increase $\alpha$ to improve the encoder's prediction accuracy. When $\alpha$ is above a threshold, we detach the prediction error loss of all modes but the one with the largest $Q$ in \cref{eq:expectation_loss} from the gradient graph to avoid mode collapse. This allows us to reduce the mode collapse under a small $\alpha$ while continue to improve the encoder on the mode probability prediction.

\subsection{Diverse Sampling from Product Latent Space}\label{sec:product_sampling}
Since $P(\mathbf{z}|\mathbf{x})$ is calculated with factor graphs, two clique modes with similar latent variables, e.g., two $\mathbf{z}$-s that only differ at one node in the clique, may have similar probabilities,causing the greedy sampling result to lose diversity. This issue is well-known in Markov random fields and we follow the simple diverse sampling scheme in \cite{batra2012diverse}. To be specific, we use a greedy algorithm to pick $\mathbf{z}$ one by one as 
\begin{equation*}
\begin{aligned}
    \mathbf{z}^{k+1} =&\; \mathop{\arg\max}\limits_{\mathbf{z}\in\mathcal{Z}} p(\mathbf{z}|x)\\
    \mathrm{s.t.} &\;\forall \mathbf{z}^i,i=1,...,k, \Delta(\mathbf{z},\mathbf{z}^i)\ge\beta,
\end{aligned}
\end{equation*}
where $\Delta$ is a distance function, for our setup, $\Delta(\mathbf{z}^1,\mathbf{z}^2)=|\{j|z_j^1\neq z_j^2\}|$, i.e., the number of nodes with different latent variables under $\mathbf{z}^1$ and $\mathbf{z}^2$. For example, $\Delta([0,1,2],[1,1,2])=1,\Delta([1,0,0],[2,1,0])=2$. 

\end{document}